%% file: main.tex
\documentclass[10pt,twocolumn,letterpaper]{article}

\usepackage[pagenumbers]{iccv} 


\definecolor{iccvblue}{rgb}{0.21,0.49,0.74}
\usepackage[pagebackref,breaklinks,colorlinks,allcolors=iccvblue]{hyperref}
\usepackage{subcaption}
\usepackage{graphicx} 
\usepackage{caption} 
\usepackage{siunitx}
\usepackage{amsthm}
\usepackage{algorithm}
\usepackage{algorithmic}
\usepackage{amsmath}
\usepackage{amssymb}
\usepackage[T1]{fontenc}
\usepackage{multirow}
\usepackage{multicol}
\usepackage{autobreak}
\usepackage{graphicx}
\usepackage{epstopdf}
\usepackage{enumitem} 
\usepackage{tikz}
\usepackage{diagbox}
\usepackage{times}  
\usepackage{helvet}  
\usepackage{courier}  
\usepackage[hyphens]{url}  
\usepackage{graphicx} 
\usepackage{hyperref}

\usepackage{multirow}       
\usepackage{booktabs}

\newtheorem{proposition}{Proposition}
\newtheorem{definition}{Definition} 
\def\paperID{2254} 
\def\confName{ICCV}

\title{A Self-Supervised Paradigm for Data-Efficient Medical Foundation Model Pre-training: $\mathcal{V}$-information Optimization Framework}

\author{
    Wenxuan Yang\textsuperscript{\rm 1},
    Hanyu Zhang\textsuperscript{\rm 1},
    Weimin Tan\textsuperscript{\rm 1},
    Yuqi Sun\textsuperscript{\rm 1},
    Bo Yan\textsuperscript{\rm 1}\\[2mm]
    \small
    \textsuperscript{\rm 1}Shanghai Key Laboratory of Intelligent Information Processing,
    School of Computer Science, Fudan University
}

\begin{document}
\maketitle
\input{sec/0_abstract}    
\input{sec/1_intro}

\input{sec/2_relatedwork}
\input{sec/3_preliminary}
\input{sec/4_method}

\input{sec/5_experiment}

\input{sec/6_conclution}
{
    \small
    \bibliographystyle{ieeenat_fullname}
    \bibliography{main}
}
\input{sec/7_suppl}

\end{document}

%% file: sec/0_abstract.tex
\begin{abstract}
Self-supervised pre-training medical foundation models on large-scale datasets demonstrate exceptional performance. Recent research challenges this common paradigm by introducing data-effective learning approaches, demonstrating that merely increasing pre-training data volume does not necessarily improve model performance. However, current methods still have unclear standards and the underlying theoretical foundation remains unknown. In this paper, as the first attempt to address this limitation, we introduce $\mathcal{V}$-information into self-supervised pre-training of foundation models to provide a theoretical foundation for sample selection. Our derivation confirms that by optimizing $\mathcal{V}$-information, sample selection can be framed as an optimization problem where choosing diverse and challenging samples enhances model performance even under limited training data. Under this guidance, we develop an optimized data-effective learning method (OptiDEL) to optimize $\mathcal{V}$-information in real-world medical domains by generating more diverse and harder samples. We compare the OptiDEL method with state-of-the-art approaches finding that OptiDEL consistently outperforms existing approaches across eight different datasets, with foundation models trained on only 5\% of the pre-training data achieving up to 6.2\% higher mIoU than those trained on the full dataset. Remarkably, OptiDEL demonstrates an average improvement of 4.7\% mIoU over competing methods while using 20× less training data.
\end{abstract}

%% file: sec/1_intro.tex
\vspace{-5pt}
\section{Introduction}
\label{sec:intro}

\begin{figure}[!t] 
  \centering
\includegraphics[width=1.0\linewidth]{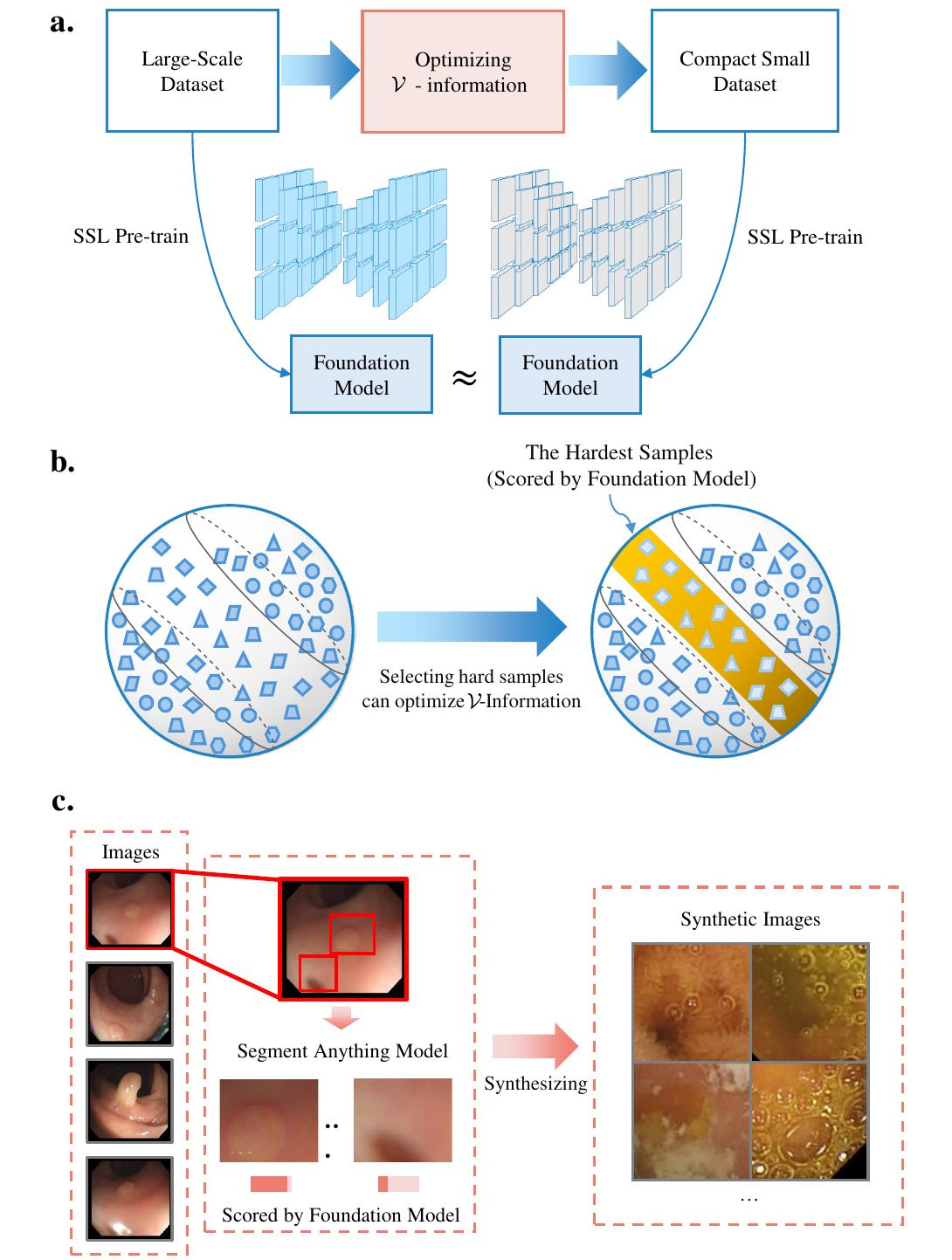} 
  \caption{Generate a $\mathcal{V}$-information-rich~\cite{Vinformation} compact dataset to self-supervised pre-train superior medical foundation models. (a) By optimizing $\mathcal{V}$-information, we can generate a smaller unlabeled dataset from a larger unlabeled dataset, with slight differences in the pre-training results of foundation models. (b) Selecting the hardest samples (that are more challenging for foundation models to classify or predict correctly) can optimize the $\mathcal{V}$-information, thereby enhancing model performance. (c) Synthesizing more diverse samples can optimize the $\mathcal{V}$-information.}
  \label{head}
  \vspace{-10pt}
\end{figure}

\begin{figure*}[t!] 
  \centering
  \includegraphics[width=1.0\textwidth]{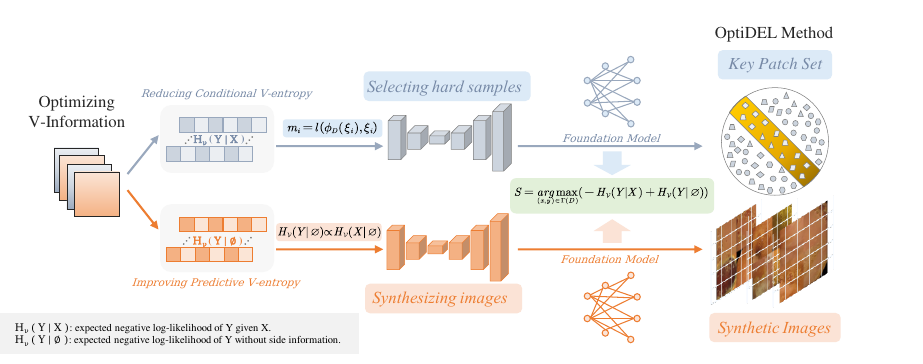} 
  \caption{The framework of our OptiDEL method to optimize the $\mathcal{V}$-information. The optimization process focuses on two key components: reducing $H_{\mathcal{V}}(Y|X)$ and enhancing $H_{\mathcal{V}}(Y|\varnothing)$. By leveraging a foundation model, we identify and select challenging patches, which helps reduce $H_{\mathcal{V}}(Y|X)$. We then combine every four of these challenging patches to create new images, thereby enhancing $H_{\mathcal{V}}(Y|\varnothing)$.}
  \label{fig:pipline}
  \vspace{-5pt}
\end{figure*}

Self-supervised pre-training medical foundation models on large-scale datasets have become standard practice, as they significantly boost model generalization capability, enabling exceptional performance with fine-tuning on limited data~\cite{kolides2023artificial}. Current approaches incorporate large amounts of low-information data into pre-training, incurring high collection and computational costs while yielding marginal performance gains. To investigate whether it is possible to achieve comparable performance with less pre-training data,~\citet{yang2024acm} introduced a benchmark for data-effective learning, exploring the significant nonlinear phenomenon exhibited by foundation models trained with different proportions of data on a given dataset. However, despite the importance of this phenomenon for self-supervised pre-training, its theoretical foundation remains unknown, and there is no clear standard for sampling strategies.

\nocite{yang2024acm}
\nocite{kolides2023artificial}

Fortunately, in the field of supervised learning, data selection methods based on hard sample selection and diversity metrics have been theoretically proven to improve data utilization efficiency and model performance. Among these, difficult sample selection techniques ~\cite{DBLP:conf/icml/MirzasoleimanBL20,DBLP:conf/iclr/TonevaSCTBG19,DBLP:conf/nips/FeldmanZ20,DBLP:conf/nips/PaulGD21}, which focus on data that models struggle to accurately predict or tend to forget during training iterations, have become effective techniques for quantifying data importance. On the other hand, diversity-aware selection paradigms that reduce data redundancy have demonstrated significant impact in optimizing dataset quality, particularly in the field of medical imaging~\cite{d2,badge,Sun_2024_Crop,yang2024acm, mahawar2024label}. However, transferring these techniques from supervised learning to self-supervised learning remains a major challenge, especially in terms of computational complexity and theoretical validation.



\nocite{DBLP:journals/corr/abs-2405-16640}
\nocite{ceccarello2018fast}

Inspired by these insights, we are the first to integrate the two key concepts of hard sample selection and data diversity into a theoretical framework for self-supervised pretraining. We observe that these two concepts align perfectly with $\mathcal{V}$-information~\cite{Vinformation} (a computationally constrained information measure that extends classical mutual information under practical operational constraints). Therefore, we introduce $\mathcal{V}$-information into the self-supervised foundation model pre-training, framing data-effective learning task as an entropy optimization problem, as shown in Fig~\ref{head}(a). Building upon this, our theoretical derivation further demonstrates that selecting harder samples (Fig~\ref{head}(b)) and increasing diversity (Fig~\ref{head}(c)) helps optimize $\mathcal{V}$-information, enabling models to potentially match or even surpass the performance of those trained on the full dataset while utilizing fewer data. Guided by this theoretical framework, we validate our approach in the medical domain by proposing the Optimized Data-Effective Learning (OptiDEL) method, which utilizes the Segment Anything Model (SAM) for information extraction and generates more diverse and challenging pre-training samples from the original data.

The key contributions can be summarized as:
\begin{itemize}
\item This paper first formulates self-supervised pre-training of medical foundation models as an optimization problem of increasing $\mathcal{V}$-information. Through theoretical derivation, we demonstrate that generating more diverse data and selecting the most challenging samples help optimize $\mathcal{V}$-information.

\item Based on the $\mathcal{V}$-information, we design the OptiDEL method which enhances the performance of foundation models by generating more diverse and harder samples through the extraction of critical data information using the SAM model.

\item We compare OptiDEL with the state-of-the-art methods across eight medical datasets. Our experimental results highlight the importance of optimizing $\mathcal{V}$-information for improving data-effective learning performance.

\end{itemize}

%% file: sec/2_relatedwork.tex
\section{Related Work}

Data distillation~\cite{lei2023comprehensive} is a technique aimed at reducing large-scale datasets by synthesizing or selectively extracting key samples to improve learning efficiency and reduce computational resource demands. 

\nocite{lei2023comprehensive}

\subsection{Traditional Data Distillation}

In the field of supervised learning, traditional data distillation can be mainly divided into two categories: optimization-based methods and coreset-based methods.

\subsubsection{Optimization-Based Methods}
Optimization-based methods have emerged as a crucial direction in data distillation, aiming to enhance dataset representation while reducing data volume through sophisticated optimization strategies. These methods can be broadly categorized into two main approaches: bi-level optimization and uni-level optimization.

Bi-level optimization methods \cite{DBLP:journals/corr/abs-1811-10959} focus on minimizing the discrepancy between surrogate models trained on synthetic data and those trained on the original dataset. This approach involves two nested optimization processes: outer optimization, which adjusts high-level model parameters to align synthetic data with the original dataset, and inner optimization, which fine-tunes low-level parameters to ensure consistent model performance across datasets. To achieve this alignment, bi-level methods rely on various metrics, including gradient matching \cite{DBLP:conf/iclr/ZhaoMB21, DBLP:conf/icml/KillamsettySRDI21}, feature matching \cite{ji2021show}, distribution matching \cite{zhang2021matching, binici2022preventing}, and training trajectory matching \cite{cazenavette2022dataset, guo2023towards}. While bi-level optimization excels in handling complex data distributions and tasks, its high computational complexity and resource consumption often lead to performance bottlenecks in practical applications.

To address the limitations of bi-level optimization, uni-level optimization methods \cite{DBLP:conf/nips/NguyenNXL21, DBLP:journals/corr/abs-2302-06755} simplify the optimization process, reducing training costs and improving efficiency. A prominent example is kernel ridge regression \cite{DBLP:conf/iclr/NguyenCL21, xu2023kernel}, which solves a regularized linear regression problem to distill data efficiently. Uni-level optimization significantly reduces computational resource consumption and demonstrates strong performance on large-scale datasets. However, its simplified approach may not fully capture the complexity of data and task-specific characteristics, limiting its applicability in certain scenarios.

\subsubsection{Coreset-Based Methods}
Coreset-based methods~\cite{DBLP:conf/icml/MirzasoleimanBL20,DBLP:conf/icml/PooladzandiDM22} focus on identifying and selecting a subset of samples that can represent the entire dataset. These methods are gaining attention for their computational efficiency and ability to retain the representativeness of the original dataset. They typically utilize metrics such as Forgetting Score~\cite{DBLP:conf/iclr/TonevaSCTBG19}, Memorization~\cite{DBLP:conf/nips/FeldmanZ20}, and EL2N Score~\cite{DBLP:conf/nips/PaulGD21} to evaluate the importance of samples. Additionally, they employ Diverse Ensembles\nocite{DBLP:conf/iclr/MedingBGW22} strategies to ensure that the selected samples comprehensively cover the diversity present in the dataset.

\nocite{DBLP:journals/corr/abs-1811-10959}
\nocite{DBLP:conf/nips/NguyenNXL21}
\nocite{DBLP:journals/corr/abs-2302-06755}

\nocite{DBLP:conf/iclr/ZhaoMB21}
\nocite{DBLP:conf/icml/KillamsettySRDI21}

\nocite{ji2021show}

\nocite{zhang2021matching}
\nocite{binici2022preventing}

\nocite{cazenavette2022dataset}
\nocite{guo2023towards}

\nocite{DBLP:conf/iclr/NguyenCL21}
\nocite{xu2023kernel}

\nocite{DBLP:conf/icml/MirzasoleimanBL20}
\nocite{DBLP:conf/icml/PooladzandiDM22}

\nocite{DBLP:conf/iclr/TonevaSCTBG19}
\nocite{DBLP:conf/nips/FeldmanZ20}
\nocite{DBLP:conf/nips/PaulGD21}
\nocite{DBLP:conf/iclr/MedingBGW22}

\subsection{Data-Effective Learning Methods}

 Unlike traditional supervised data distillation,~\citet{yang2024acm} have proposed a comprehensive benchmark specifically designed to evaluate data-effective learning in the medical field. This research focuses on self-supervised data-effective learning aimed at efficiently training foundation models. The benchmark includes a dataset with millions of data samples from 31 medical centers (DataDEL), a baseline method for comparison (MedDEL), and a new evaluation metric (NormDEL) to objectively measure the performance of data-effective learning. This benchmark lays the foundation for the pre-training and theoretical validation of foundation models. Additionally, ~\cite{SAS} leveraged contrastive learning to identify and select the most important data subset, which is then used to train the foundation model.

%% file: sec/3_preliminary.tex
\section{Preliminary}
\subsection{Data-effective Learning}
The goal of data-effective learning is to generate a small-scale dataset $S=\left\{ X|X=x_i \right\} _{i=1}^{|S|}$ from a large number of unlabeled data samples $D=\left\{ \hat{X}|\hat{X}=\hat{x}_j \right\} _{j=1}^{|D|}$, where $|S|\ll |D|$. The essence of this approach is to ensure that the pre-training performance of the foundation model on the small dataset $S$ is comparable to its performance on the large dataset $D$ within an acceptable error range. 


\nocite{He_2022_CVPR}
\nocite{ViT}

\subsection{Predictive V-information}

$\mathcal{V}$-information is a variational extension to classic mutual information
which attempts to capture usable information and is effective for structure learning and fair representation learning. As defined in \cite{Vinformation}:

\begin{definition}
Let $X$, $Y$ be two random variables taking values in $\mathcal{X} \times \mathcal{Y}$, and let $\mathcal{V}$ represent a predictive family. The predictive $\mathcal{V}$-information from $X$ to $Y$ is defined as:
\begin{equation}
    I_{\mathcal{V}}(X \to Y) = -H_{\mathcal{V}}(Y|X) + H_{\mathcal{V}}(Y|\varnothing),
\end{equation}
where:
\begin{itemize}
    \item The first term $H_{\mathcal{V}}(Y|X)$ is the conditional $\mathcal{V}$-entropy:
    \begin{equation}
        H_{\mathcal{V}}(Y|X) = \inf_{f \in \mathcal{V}} \mathbb{E}_{x,y \sim X,Y} \left[ -\log f[x](y) \right];
    \end{equation}
    \item The second term $H_{\mathcal{V}}(Y|\varnothing)$ is the predictive $\mathcal{V}$-entropy:
    \begin{equation}
        H_{\mathcal{V}}(Y|\varnothing) = \inf_{f \in \mathcal{V}} \mathbb{E}_{y \sim Y} \left[ -\log f[\varnothing](y) \right].
    \end{equation}
\end{itemize}
Here, $f[x](y)$ is a probability measure on $y \in \mathcal{Y}$ given side information $x$, and $f[\varnothing](y)$ denotes the measure with no side information.
\end{definition}



%% file: sec/4_method.tex
\section{Methodology}
~\citet{yang2024acm} have provided a benchmark in the field of medical date-effective learning. However, their approach is relatively fundamental, leaving substantial room for further exploration. In the following sections, we demonstrate that the data-effective learning task can be framed as an optimization problem by optimizing $\mathcal{V}$-information in the self-supervised learning task. Building on this foundation and incorporating principles from information theory, we will also detail our OptiDEL algorithm.

\subsection{V-information based SSL Framework}
\citet{vusable} explore the application of $\mathcal{V}$-information in a classification task, where $\mathcal{V}$ represents a neural network model, $X$ denotes the text input, and $Y$ is the corresponding gold label. Building on this concept, \citet{Sun_2024_Crop} proposed a supervised data distillation method based on $\mathcal{V}$-information for natural image classification. Specifically, the goal is to select a smaller subset $S$ from the dataset $D$ through some algorithm $\Gamma$ that optimizes the $\mathcal{V}$-information from the input image $X$ to the gold label $Y$. This process can be written as:
\begin{align}
S=\mathop {arg\max} \limits_{(X,Y)\in \Gamma(D)}I_{\mathcal{V}}\left( X\rightarrow Y \right).
\label{eq:S}
\end{align}

We apply this approach to pre-train a self-supervised foundation model on the selected subset $S$ and validate its effectiveness on downstream tasks. Here, $X$ represents the input image, and the gold label $Y$ is the corresponding self-supervised target. Denoting the self-supervised foundation model as $\mathcal{V}$, we can further express equation \eqref{eq:S} as:
\begin{align}
S&= \mathop {arg\max} \limits_{(X,Y)\in \Gamma(D)} ( \underbrace{-H_{\mathcal{V}}(Y|X)}_{\text{conditional $\mathcal{V}$-entropy}} + \underbrace{H_{\mathcal{V}}(Y|\varnothing))}_{\text{predictive $\mathcal{V}$-entropy}}. \label{eq:S2}
\end{align}

\nocite{Sun_2024_Crop}
\nocite{Vinformation}

Our optimization strategy focuses on increasing the predictive $\mathcal{V}$-information on the selected subset $S$. This involves reducing the first term $H_{\mathcal{V}}(Y|X)$ and enhancing the second term $H_{\mathcal{V}}(Y|\varnothing)$. In the following section, we demonstrate that we can reduce the first term by selecting hard examples from $X$ and enhance the second term by improving the diversity of $Y$, thus improving the performance of the foundation model in downstream tasks.


\subsection{Optimizing V-Information in SSL task}
\label{sec:4.2}

\subsubsection{Reducing Conditional Entropy via Hard Samples}
\vspace{2pt}
\noindent
\textbf{Approximation of the conditional $\mathcal{V}$-entropy $\mathbf{H_{\mathcal{V}}(Y|X)}$. }First, we need to estimate the conditional entropy term $H_{\mathcal{V}}(Y|X)$, which represents the uncertainty involved in predicting $Y$ given $X$. 
\citet{vusable} proposed a method to estimate this term in supervised learning by leveraging a held-out test set. 
However, in self-supervised foundation model pre-training, the absence of labels and the challenge of constructing a distributionally-aligned held-out set make existing methods inapplicable for evaluating $H_{\mathcal{V}}(Y|X)$. To address the above problems, we approximate $H_{\mathcal{V}}(Y|X)$ on downstream tasks, leveraging the inherent correlation between downstream tasks and the pre-training data. 

 Consider selecting a subset $D'$ from the original dataset $D$. Let $f_{D'}$ denote the foundation models pre-trained on $D'$. We estimate $H_{\mathcal{V}}(Y|X)$ on a downstream dataset $D_{\text{d}} = \{(x_i, y_i)\}_{i=1}^n$, where $x_i$ is the input and $y_i$ is the gold label. The conditional entropy for $D'$ is:
\begin{equation}
    H_{\mathcal{V}}^{D'}(Y|X) = -\frac{1}{n} \sum_{(x_i, y_i) \in D_{\text{d}}} \log f_{D'}[x_i](y_i).
\end{equation}

Since the conditional $\mathcal{V}$-information of the model pre-trained on the entire dataset $D$ is fixed, the data selection objective under a pruning ratio $\alpha$ can be formulated as:
\begin{align}
D' &= \mathop{\arg\max}\limits_{ |D'|=\alpha|D|}  H_{\mathcal{V}}^{D'}(Y|X) \\
   &\approx  \mathop{\arg\max}\limits_{ |D'|=\alpha|D|}  \sum_{x_i, y_i \in D_{\text{d}}}  \log f_{D'}[x_i](y_i).
\label{eq:sub}
\end{align}
Here, $f[x](y)$ represents the log probability assigned by the model $\mathcal{V}$ to the gold label $y$ given the input $x$ and it is highly correlated with model performance. 

\begin{figure}[t!]
  \centering
    \includegraphics[width=0.65\columnwidth]{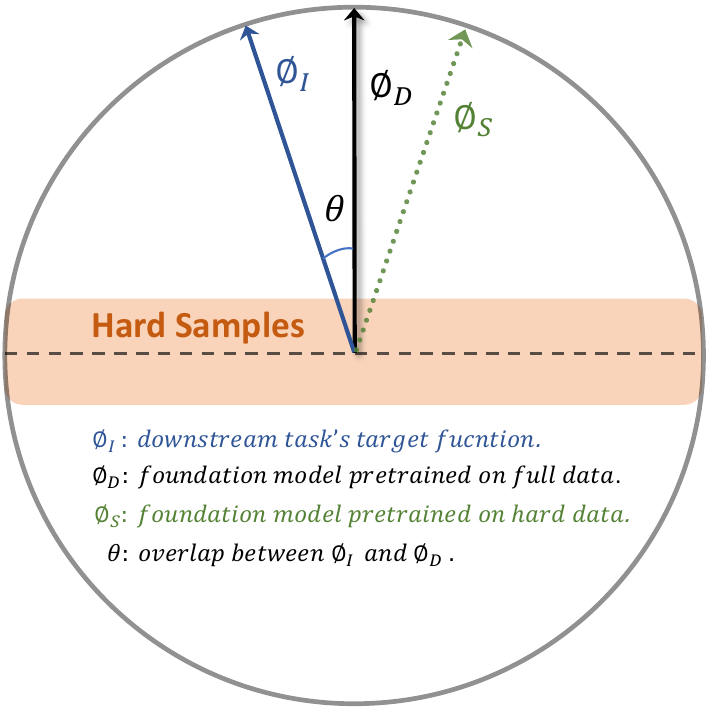}
    \caption{Toy example illustrating the impact of hard sample selection on downstream task alignment: the target vector $\boldsymbol{\phi_I}$ represents the downstream task’s objective, the probe vector $\boldsymbol{\phi_D}$ is pre-trained on full Gaussian-distributed data and imperfectly aligns with $\boldsymbol{\phi_I}$ (with the angle $\theta$ quantifying the mismatch), and hard samples are selected near $\boldsymbol{\phi_D}$’s decision boundary, on which a new predictor $\boldsymbol{\phi_S}$ is trained from scratch.}
    \label{fig:toy_example}
\end{figure}

\vspace{2pt}
\noindent
\textbf{Selecting hard samples helps to reduce the conditional $\mathcal{V}$-entropy $\mathbf{H_{\mathcal{V}}(Y|X)}$. }The optimization objective in Equation \eqref{eq:sub} raises a fundamental question: How can we strategically select examples for pre-training self-supervised foundation models to enhance downstream task performance under a fixed data selection ratio? This aligns with difficulty-based data selection, where incorporating hard examples (i.e., those more challenging for models) has shown benefits in upstream tasks~\cite{DBLP:conf/iclr/TonevaSCTBG19, DBLP:conf/nips/FeldmanZ20, DBLP:conf/nips/PaulGD21}. Our work extends this by emphasizing the advantages of hard example selection for downstream tasks.  We propose the following proposition to illustrate this finding formally:
\begin{proposition}
\label{pro}
Selecting the hardest samples $D_{\text{hard}}$ from the dataset $D$ enhances model performance on downstream tasks under a fixed selection ratio and large total data volumes. Formally, this can be expressed as:
\begin{equation}
\sum_{x_i, y_i \in D_{\text{d}}}  \left(\log f_{D_\text{hard}}^{}[x_i](y_i)-\log f_{D_\text{random}}[x_i](y_i)\right )>0
\end{equation}
\end{proposition}

This proposition demonstrates that the selection of hard examples can effectively improve the term in equation \eqref{eq:sub}, thus increase the conditional entropy term $H_{\mathcal{V}}(Y|X)$.

To confirm Proposition 1, we extend \citet{sorscher}'s numerical experiments using a toy example to examine the impact of hard example selection on downstream tasks. (1) The task involves predicting a target vector $\mathbf{\phi_I}$ sampled from $\mathrm{Unif}(\mathbf{S}^{N-1}(\sqrt{N}))$, which represents the downstream task's target function; (2) The training dataset $D$ consists of vectors $\mathbf{x}^\mu$ randomly sampled from an $N$-dimensional Gaussian distribution, with corresponding labels $y^\mu = \text{sign}(\mathbf{\phi_I} \cdot \mathbf{x})$; (3) We pre-train a probe vector $\mathbf{\phi_{D}}$ on dataset $D$, which can be considered as the foundation model's target function though it may imperfectly match the target $\mathbf{\phi_I}$ due to the domain gap between pre-training and downstream tasks, and hard examples are identified as samples with the smallest margin $m^\mu = \mathbf{\phi_{D}} \cdot (y^\mu \mathbf{x}^\mu)$, where a smaller margin indicates that the vector is closer to the decision boundary, i.e., the region where the label changes and fluctuates; (4) From these hard examples, we select a subset $S$ and pre-train a predictor vector $\mathbf{\phi_S}$ from scratch.


We assess the impact of hard example selection by quantifying the correlation $ R $ between $\mathbf{\phi_S}$ and $\mathbf{\phi_I}$.  We demonstrate that $ R $ under pruning ratio $f$ can be derived by solving the following system of equations:
\begin{equation}
\left\{
\begin{aligned}
& \frac{R - \rho \cos\theta}{\sin^2\theta} = \frac{cf}{\pi \Psi} \langle \int_{-\infty}^{m} dx \, \exp\left(-\frac{G(x,z)}{2\Psi^2}\right) (m - x) \rangle_z \\
& \frac{\sin^2\theta-\rho^2 - R^2 + 2\rho R \cos\theta}{\sin^2\theta} = 2cf \langle \int_{-\infty}^{\kappa} (m - x)^2 dx \\
& \frac{e^{-\frac{(x-\rho z)^2}{2(1-\rho^2)}}}{\sqrt{2\pi(1-\rho^2)}}   
H\left(\frac{F(x,z)}{\sqrt{1-\rho^2}\Psi}\right) \rangle_z \\
& \frac{\rho - R\cos\theta}{\sin^2\theta} = 2cf \langle \int_{-\infty}^{m} dx \, \frac{e^{-\frac{(x-\rho z)^2}{2(1-\rho^2)}}}{\sqrt{2\pi(1-\rho^2)}} (m - x) \\ 
& H\left(\frac{F(x,z)}{\sqrt{1-\rho^2}\Psi}\right) \frac{z - \rho x}{1-\rho^2} \rangle_z
\end{aligned}
\right.
\label{eq:hard}
\end{equation}
where the auxiliary functions are defined as:
\begin{equation}
\left\{
\begin{aligned}
& \Psi = \sqrt{\sin^2\theta - R^2 - \rho^2 + 2\rho R \cos\theta} \\
& F(x,z) = z(\rho R - \cos\theta) - x(R - \rho\cos\theta) \\
& G(x,z) = z^{2}(\rho^{2} + \cos^{2}\theta - 2\rho R\cos\theta) \\
& \quad + 2xz(R\cos\theta - \rho) + x^{2}\sin^{2}\theta \\
& \rho = \mathbf{\phi_{D}} \cdot \mathbf{\phi_{S}}
\end{aligned}
\right.
\end{equation}

For further theoretical analysis, detailed additional parameter descriptions in the equations, and information on the numerical setup, please refer to the Appendix. And a comprehensive discussion is provided in \cref{foundation}.

\subsubsection{Enhancing Predictive Entropy via Diversity}

The predictive entropy term $H_{\mathcal{V}}(Y|\varnothing)$ represents the minimum expected negative log-likelihood for predicting $Y$. In self-supervised learning tasks, $Y$ is often closely related to the input data $X$. For instance, in the MAE architecture, the input $X$ is a masked version of the original image, while the label $Y$ corresponds to the original pixel values of the masked regions. Such inherent coupling implies a positive correlation between the information content of $X$ and $Y$, leading to:
\begin{equation}
    H_{\mathcal{V}}(Y|\varnothing) \propto H_{\mathcal{V}}(X|\varnothing).
\end{equation}
Consequently, optimizing $H_{\mathcal{V}}(Y|\varnothing)$ can be directly accomplished by enhancing $H_{\mathcal{V}}(X|\varnothing)$, which can be achieved by increasing the diversity of the data.

\subsection{Details of OptiDEL Algorithm}

In this section, we provide the details of the proposed OptiDEL algorithm inspired by optimizing $\mathcal{V}$-information. A simplified version of the OptiDEL algorithm is illustrated through the pseudocode presented in Algorithm~\ref{alg:algorithm}. For our self-supervised foundation model architecture, we employ the Masked Auto Encoder (MAE)~\cite{He_2022_CVPR} and SimCLR ~\cite{SimCLR} frameworks—both widely adopted in medical applications due to their demonstrated capability to extract powerful representations from extensive unlabeled datasets.

\subsubsection{Reducing Conditional Entropy in OptiDEL}
\textbf{Using SAM to Extract Key Information.} We extract several patches from the original image set $D$ and use them for subsequent pre-training of the foundation model to improve the diversity of the data.
Current method~\cite{Sun_2024_Crop} involves random patch cropping, which inherently introduces a degree of randomness. Ideally, we propose leveraging the segment anything model (SAM)~\cite{SAM} to extract potential lesions in medical images proactively. The entire process can be formulated as:
\begin{align}
\xi _{i,*}=S\left( \hat{x}_i \right) =\left\{ S\left( \hat{x}_{i,1} \right),\cdots ,S\left( \hat{x}_{i,k} \right) \right\}.
\end{align}
Subsequently, we apply this operation to all the original images, resulting in $S=\lbrace \xi_{i,*} \rbrace_{i=1}^{|D|}=\lbrace \xi_i\rbrace_{i=1}^{N\times |D|} $, where $\xi _{i,k}$ represents the $k^{th}$ patch extracted from the $i^{th}$ image.

\nocite{SAM}

\vspace{2pt}
\noindent
\textbf{Patch selection as hard samples to reduce the conditional $\mathcal{V}$-entropy. }The MAE model can be considered a function that maps a masked input back to its original version, which can perfectly reconstruct patches in the ideal situation. However, in practice, some patches are difficult to reconstruct while others are relatively easier, leading to variable reconstruction errors in actual trained foundation model $\phi_D$. Inspired by this, we define the margin $ m_{i}$ for the patch $\xi_{i}$ based on the reconstruction loss from the pre-trained MAE model $\phi_D$:
\begin{align}
     m_{i} = l\left(\phi_D(\xi_{i}),\xi_{i}\right).
\end{align}

We classify patches with higher margins as harder examples, while those with smaller margins are considered easier. In the following sections, we show that prioritizing patches with the largest margins for subsequent synthesis not only saves computation time but also maintains or even enhances the model's performance on downstream tasks.

\nocite{sorscher}
\begin{algorithm}[t]
\caption{Pseudo Code for OptiDEL}
\label{alg:algorithm}
\textbf{Input}: A dataset $D$ and a pre-trained self-supervised foundation model $\phi_D$ (here we take MAE as an example). \\
\textbf{Parameter}: $N, K, M$
\begin{algorithmic}[1] 
\FOR{$\hat{x_i} \in D$}
    \STATE Utilize SAM to crop $\hat{x_i}$ into $N$ patches $\lbrace \xi_{i}\rbrace_{i=1}^{N}$.
\ENDFOR
\STATE Get patches set $\lbrace \xi_{i}\rbrace_{i=1}^{N \times |D|}$.
\FOR{$i=1$ \textbf{to} $N \times |D|$}
    \STATE Calculate reconstruction loss of MAE model as margin $m_{i} = l\left(\phi_D(\xi_{i}), \xi_{i}\right)$ of patch $\xi_{i}$.
\ENDFOR
\STATE Select top-$K$ margin patches $\lbrace \xi^{'}_i \rbrace_{i=1}^{K}$.
\FOR{$i=1$ \textbf{to} $K$}
    \IF{$i \mod M == 0$}
        \STATE Synthesize $M$ patches $\lbrace \xi^{'}_j \rbrace_{j=i+1-M}^{i}$ into a new image $X_j$.
    \ENDIF
\ENDFOR
\end{algorithmic}
\textbf{Output}: Synthesized dataset $S = \lbrace X \mid X = X_j \rbrace$ and the final foundation model $\phi_S$ trained from scratch on $S$.
\vspace{5pt}
\end{algorithm}

\subsubsection{Enhancing Predictive Entropy in OptiDEL}
\textbf{Patch synthesis to enhance diversity to enhance predictive $\mathcal{V}$-entropy.} After selecting image patches as hard examples, we synthesize these patches into larger images. Typically, the size of each patch is smaller than the final image we aim to create. Thus, from our selected set of hard patches denoted as $\lbrace \xi^{'}_i\rbrace_{i=1}^{K}$, we synthesize every $M$ patch into a new image in order of hardness, from the hardest to the simplest, to compile them into a larger, cohesive image $x_j$ and keep the pixels of the synthesized image consistent with those of the original dataset.

\vspace{2pt}
\noindent
After completing the aforementioned steps, we conduct self-supervised pre-training with MAE and SimCLR on the generated dataset $S$ to obtain our foundation model.

\vspace{10pt}
\noindent
\textbf{Summary. }\textit{We propose a patch extraction mechanism that identifies and selects the most challenging regions from source images for synthesis. This dual-objective framework facilitates both the integration of heterogeneous patches within individual synthesized images and the concentration of hard examples. Through comprehensive theoretical analysis, we demonstrate that our approach significantly enhances the $\mathcal{V}$-information of the selected subset, thereby improving the overall quality of the synthesized data.}


~\nocite{DBLP:conf/iclr/TonevaSCTBG19}
~\nocite{DBLP:conf/nips/FeldmanZ20}
~\nocite{DBLP:conf/nips/PaulGD21}
~\nocite{sorscher}
~\nocite{ccs}
~\nocite{d2}

%% file: sec/5_experiment.tex
\begin{table*}[ht!]
    \centering
    \caption{Performance comparison of OptiDEL with state-of-the-art data-effective learning methods in MAE (ViT-Base) structure. To evaluate the performance of our OptiDEL method, we pre-train MAE (ViT-Base) using 5\%, 10\%, 25\%, and 50\% of the total data volume and test them on eight downstream datasets using EL2N, RDED, MedDEL, and OptiDEL. The downstream segmentation is conducted using DPT (ViT-Base). OptiDEL consistently outperforms the other methods across all datasets, demonstrating its stability and efficiency.}
    \renewcommand{\arraystretch}{1.35}  
    \setlength{\tabcolsep}{4pt}  
    \resizebox{\textwidth}{!}{  
        \begin{tabular}{l| c | c c c c | c c c c | c c c c | c c c c}
            \toprule
            \multirow{2}{*}{\diagbox{\textbf{Dataset}}{\textbf{mIoU}}} 
            & \textbf{All Data} 
            & \multicolumn{4}{c}{\textbf{5\% storage volume}} 
            & \multicolumn{4}{|c}{\textbf{10\% storage volume}} 
            & \multicolumn{4}{|c}{\textbf{25\% storage volume}} 
            & \multicolumn{4}{|c}{\textbf{50\% storage volume}} \\
            \cmidrule(lr){2-2} \cmidrule(lr){3-6} \cmidrule(lr){7-10} \cmidrule(lr){11-14} \cmidrule(lr){15-18}
            & \textbf{Baseline} 
            & \textbf{EL2N} & \textbf{RDER} & \textbf{MedDEL} & \textbf{OptiDEL} 
            & \textbf{EL2N} & \textbf{RDER} & \textbf{MedDEL} & \textbf{OptiDEL} 
            & \textbf{EL2N} & \textbf{RDER} & \textbf{MedDEL} & \textbf{OptiDEL} 
            & \textbf{EL2N} & \textbf{RDER} & \textbf{MedDEL} & \textbf{OptiDEL} \\
            \midrule
            Kvasir-Instrument & 77.78 & 83.25 & 83.51 & 79.76 & \textbf{83.99} & 83.75 & 83.54 & 80.11 & \textbf{84.26} & 76.68 & 83.19 & 80.44 & \textbf{83.78} & 76.72 & 79.76 & 79.53 & \textbf{83.38} \\
            Kvasir-SEG & 60.81 & 60.69 & 65.41 & 62.20 & \textbf{65.97} & 64.72 & 65.39 & 63.15. & \textbf{66.93} & 63.93 & 64.80 & 60.57 & \textbf{67.14} & 59.90 & 62.80 & 62.46 & \textbf{65.88} \\
            ImageCLEFmed & 59.64 & 60.43 & 60.32 & 57.08 & \textbf{62.04} & 59.95 & 59.88 & 56.45 & \textbf{64.21} & 59.93 & 59.91 & 56.29 & \textbf{66.80} & 59.45 & 61.43 & 56.53 & \textbf{62.04} \\
            ETIS & 51.64 & 53.26 & 55.16 & 53.05 & \textbf{56.48} & 52.84 & 56.80 & 54.40 & \textbf{57.87} & 53.40 & 55.51 & 57.58 & \textbf{64.23} & 50.43 & 51.64 & 52.13 & \textbf{56.13} \\
            PolypGen2021 & 57.11 & 54.82 & 55.00 & 54.64 & \textbf{57.01} & 56.35 & 57.11 & 53.60 & \textbf{58.56} & 55.60 & 55.14 & 54.34 & \textbf{58.44} & 57.32 & 53.91 & 55.08 & \textbf{56.82} \\
            CVC-300 & 83.26 & 83.99 & 84.12 & 83.84 & \textbf{86.92} & 85.22 & 86.13 & 83.66 & \textbf{87.90} & 85.55 & 83.30 & 84.51 & \textbf{87.22} & 85.48 & 85.90 & 85.22 & \textbf{87.30} \\
            CVC-ClinicDB & 82.78 & 79.97 & 80.19 & 79.25 & \textbf{82.49} & 81.11 & 81.29 & 80.98 & \textbf{83.32} & 82.49 & 82.77 & 81.92 & \textbf{83.93} & 81.74 & 82.01 & 81.03 & \textbf{83.21} \\
            CVC-ColonDB & 76.74 & 76.31 & 75.19 & 75.72 & \textbf{77.84} & \textbf{77.77} & 74.99 & 75.79 & 77.18 & 76.69 & 76.33 & 75.74 & \textbf{78.59} & 76.22 & 75.39 & 77.42 & \textbf{77.81} \\
            \bottomrule
        \end{tabular}
    }  
        \label{app_tab3}
        \vspace{10pt}
\end{table*}

\begin{table*}[!ht]
    \caption{Performance comparison of OptiDEL with state-of-the-art data-effective learning methods in SimCLR (ResNet-50) structure. To evaluate the effectiveness of our OptiDEL method, we pre-train SimCLR (ResNet-50) models using 5\%, 10\%, 25\%, and 50\% of the total data volume and test them on eight downstream medical image segmentation datasets using MedDEL, SAS, and our proposed OptiDEL. The downstream segmentation is conducted using DeepLab-V3 (ResNet-50). OptiDEL consistently outperforms the other methods across all datasets, demonstrating its stability and efficiency even with limited training data.}
    \centering
    \renewcommand{\arraystretch}{1.35}  
    \setlength{\tabcolsep}{7pt}  
    \resizebox{\textwidth}{!}{  
        \scriptsize 
        \begin{tabular}{l|c|ccc|ccc|ccc|ccc}
        \toprule
        \multirow{2}{*}{\diagbox{\textbf{Dataset}}{\textbf{mIoU}}} & \multicolumn{1}{c}{\textbf{All Data}} & \multicolumn{3}{|c}{\textbf{5\% storage volume}} & \multicolumn{3}{|c}{\textbf{10\% storage volume}} & \multicolumn{3}{|c}{\textbf{25\% storage volume}} & \multicolumn{3}{|c}{\textbf{50\% storage volume}} \\
        \cmidrule(lr){2-2} \cmidrule(lr){3-5} \cmidrule(lr){6-8} \cmidrule(lr){9-11} \cmidrule(lr){12-14}
         & \textbf{Baseline} & \textbf{MedDEL} & \textbf{SAS} & \textbf{OptIDEL} & \textbf{MedDEL} & \textbf{SAS} & \textbf{OptIDEL} & \textbf{MedDEL} & \textbf{SAS} & \textbf{OptIDEL} & \textbf{MedDEL} & \textbf{SAS} & \textbf{OptIDEL} \\
        \midrule
        Kvasir-Instrument & 90.46 & 85.17 & 83.09 & \textbf{90.87} & 89.25 & 89.48 & \textbf{91.16} & 89.07 & 89.61 & \textbf{91.43} & 90.34 & 90.29 & \textbf{90.81} \\
        Kvasir-SEG & 79.07 & 66.63 & 67.98 & \textbf{79.27} & 78.54 & 78.35 & \textbf{79.91} & 76.25 & 78.06 & \textbf{79.94} & 79.29 & \textbf{79.65} & 79.42 \\
        ImageCLEFmed & 77.52 & 75.23 & 75.98 & \textbf{77.92} & 76.87 & 76.56 & \textbf{78.13} & 77.67 & 77.69 & \textbf{77.98} & 77.87 & 77.95 & \textbf{78.32} \\
        ETIS & 75.17 & 74.15 & 74.73 & \textbf{76.29} & 75.42 & 75.09 & \textbf{77.13} & 76.34 & 75.46 & \textbf{76.82} & 76.29 & 76.44 & \textbf{76.50} \\
        PolyGen2021 & 70.23 & 68.46 & 68.32 & \textbf{70.00} & 68.61 & 68.79 & \textbf{70.43} & 69.44 & 69.88 & \textbf{71.59} & 69.78 & 69.97 & \textbf{70.66} \\
        CVC-300 & 87.08 & 84.81 & 83.11 & \textbf{87.37} & 86.19 & 86.61 & \textbf{87.16} & 87.07 & 86.94 & \textbf{87.65} & 87.34 & 87.38 & \textbf{87.63} \\
        CVC-ClinicDB & 83.48 & 81.73 & 81.84 & \textbf{83.19} & 82.20 & 81.77 & \textbf{83.62} & 82.85 & 82.34 & \textbf{83.94} & 83.29 & 83.10 & \textbf{83.79} \\
        CVC-ColonDB & 79.67 & 77.33 & 76.44 & \textbf{79.95} & 78.23 & 78.94 & \textbf{80.38} & 78.51 & 78.89 & \textbf{80.09} & 79.33 & 79.25 & \textbf{80.21} \\
        \bottomrule
    \end{tabular}
    }
\label{tab:SOTA2}
\vspace{10pt}
\end{table*}

\section{Experiment}
In this section, we will explore our proposed property of selecting difficult samples and synthesizing diverse samples to enhance $\mathcal{V}$-information, and demonstrate the performance of the OptiDEL method across eight downstream datasets under the guidance of this property.
\subsection{Settings}
\textbf{Datasets.} During the pre-training phase, we train the foundation model on two extensive, unlabeled datasets: LDPolypVideo~\cite{LDVideo} and Hyper-Kvasir~\cite{HyperKvasir}, which together comprise a total of 2,857,772 images. This comprehensive pre-training enables the model to learn robust features from a diverse range of images. For validation of downstream tasks, we utilize the eight segmentation datasets outlined in the DataDEL~\cite{yang2024acm}, ensuring a thorough evaluation across various segmentation challenges.

\nocite{LDVideo}
\nocite{HyperKvasir}

\vspace{2pt}
\noindent
\textbf{Model architecture. }We utilize MAE-ViT-Base and SimCLR (ResNet-50) for upstream pre-training and evaluate the effectiveness of our method on downstream tasks using Dense Prediction Transformer (DPT)~\cite{DPT} and DeepLabV3.

\vspace{2pt}
\noindent
\textbf{Hyper-parameter configuration. }In the OptiDEL task, new images are synthesized by merging every 4 patches, with experiments conducted on 5\%, 10\%, 25\%, and 50\% of the total data volume.

\vspace{2pt}
\noindent
\textbf{Baselines. }We consider data selection methods that correlate with our OptiDEL and scale to large, high-resolution datasets, particularly for self-supervised learning.
\begin{itemize}
    \item \textbf{EL2N}~\cite{DBLP:conf/nips/PaulGD21}: EL2N is a difficulty-based sample selection methodology that quantifies sample complexity through the average L2 norm of error vectors. While originally designed as a supervised data pruning technique, we adapt it to our self-supervised learning framework by initially training the foundation model over a 10-epoch period.
    \item \textbf{RDER}~\cite{Sun_2024_Crop}: RDER is a recent supervised dataset distillation method that emphasizes both the diversity and realism of the data. We apply this method to the field of self-supervised learning by using the MAE model.
    \item
    \textbf{SAS}~\cite{SAS}: SAS leverages contrastive learning to identify and select the most important subset, which is then used to train the foundation model.
    \item 
    \textbf{MedDEL}~\cite{yang2024acm}: MedDEL is a recently proposed state-of-the-art (SOTA) method for self-supervised pre-training of medical foundation models.
\end{itemize}

\nocite{DPT}

\subsection{Comparison with SOTA DEL Methods}
To further quantify the performance of our proposed OptiDEL method, we pre-train foundation models with 5\%, 10\%, 25\%, and 50\% of the total data volume. For MAE (ViT-Base), we test the performance against EL2N, RDER, and MedDEL on eight downstream datasets, utilizing DPT and as downstream frameworks (see ~\cref{app_tab3} for detailed results). In the case of SimCLR (ResNet-50), we conduct experiments against MedDEL and SAS methods, as shown in ~\cref{tab:SOTA2}. The results show that, with the same storage volume of pre-training data, the OptiDEL method consistently outperforms other methods across all downstream datasets, which highlights the robustness and effectiveness of the OptiDEL method.

Compared to competing methods, OptiDEL delivers superior performance across different model architectures and consistently achieves the highest mIoU scores on most datasets regardless of storage volume. In the MAE (ViT-Base) with DPT architecture, OptiDEL outperforms EL2N, RDER, and MedDEL in 31 out of 32 testing scenarios across the eight medical datasets. These significant performance differences underscore the effectiveness of OptiDEL's approach to data-effective learning, demonstrating that its information-rich selection process provides stable and consistent improvements across different datasets and storage constraints.

Notably, the OptiDEL algorithm demonstrates exceptional performance in extremely data-limited environments (only 5\%-25\% of original data), even surpassing results achieved with 100\% of the training data. This is particularly significant for the medical imaging field where data acquisition is often restricted. However, higher compression ratios have limited practical value, as the computational cost of data screening may be comparable to the computing power required to directly train with that portion of data.




\begin{table}[t!]
    \centering
    \caption{Analysis of the hard sample selection strategy across different foundation models using the DPT architecture. Our results indicate that OptiDEL (ImageNet), despite showing marginally lower performance than its medical dataset-pretrained counterpart, maintains impressive effectiveness and frequently surpasses the performance of models trained on the complete dataset.}
    \vspace{5pt}
    \tiny  
    \renewcommand{\arraystretch}{1.4}
    \setlength{\tabcolsep}{4pt}
    \begin{tabular}{l|c|cccc|cccc}
        \toprule
        \multirow{2}{*}{\diagbox{\textbf{Dataset}}{\textbf{mIoU}}} & \multicolumn{1}{c|}{\textbf{All Data}} & \multicolumn{4}{c|}{\textbf{OptiDEL (ImageNet)}} & \multicolumn{4}{c}{\textbf{OptiDEL (Medical)}} \\
        \cmidrule(lr){2-2} \cmidrule(lr){3-6} \cmidrule(lr){7-10}
        & \textbf{Baseline} & \textbf{5\%} & \textbf{10\%} & \textbf{25\%} & \textbf{50\%} & \textbf{5\%} & \textbf{10\%} & \textbf{25\%} & \textbf{50\%} \\
        \midrule
        Kvasir-Instrument & 77.78 & 82.65 & 83.58 & 82.98 & 82.92 & \textbf{83.99} & \textbf{84.26} & \textbf{83.78} & \textbf{83.38} \\
        Kvasir-SEG & 60.81 & 65.49 & \textbf{67.14} & 64.99 & 64.82 & \textbf{65.97} & 66.93 & \textbf{67.14} & \textbf{65.88} \\
        ImageCLEFmed & 59.64 & 60.98 & 63.75 & 60.82 & 61.51 & \textbf{62.04} & \textbf{64.21} & \textbf{66.80} & \textbf{62.04} \\
        ETIS & 51.64 & 55.24 & 57.52 & 57.02 & 52.28 & \textbf{56.48} & \textbf{57.87} & \textbf{64.23} & \textbf{56.13} \\
        PolypGen2021 & 57.11 & 56.38 & 58.05 & 56.88 & \textbf{59.65} & \textbf{57.01} & \textbf{58.56} & \textbf{58.44} & 56.82 \\
        CVC-300 & 83.26 & 84.21 & 87.15 & 86.17 & 85.90 & \textbf{86.92} & \textbf{87.90} & \textbf{87.22} & \textbf{87.30} \\
        CVC-ClinicDB & 82.78 & 82.08 & 82.70 & 82.98 & 83.03 & \textbf{82.49} & \textbf{83.32} & \textbf{83.93} & \textbf{83.21} \\
        CVC-ColonDB & 76.74 & 77.38 & 76.93 & 77.33 & 77.24 & \textbf{77.84} & \textbf{77.18} & \textbf{78.59} & \textbf{77.81} \\
        \midrule
        \textbf{Avg. Improvement} & - & +2.42 & +4.36 & +3.23 & +3.00 & \textbf{+3.51} & \textbf{+4.63} & \textbf{+6.30} & \textbf{+3.43} \\
        \bottomrule
    \end{tabular}
    \label{tab:optidel_comparison}
    \vspace{7pt}
\end{table}

\subsection{Hard Sample Selection across Models}
\label{foundation}

The hard example selection strategy implemented in OptiDEL is predicated upon a foundation model that has been pre-trained on medical datasets.  Our approach demonstrates remarkable flexibility in leveraging the pre-trained parameters, as it does not necessitate a stringent alignment between the pre-trained model and the specific downstream task domain. To validate this characteristic, we conduct research through both theoretical analysis and experiments validation on real-world datasets.

\begin{figure}[t]
    \begin{subfigure}[b]{0.49\columnwidth}
        \centering
        \includegraphics[width=\textwidth]{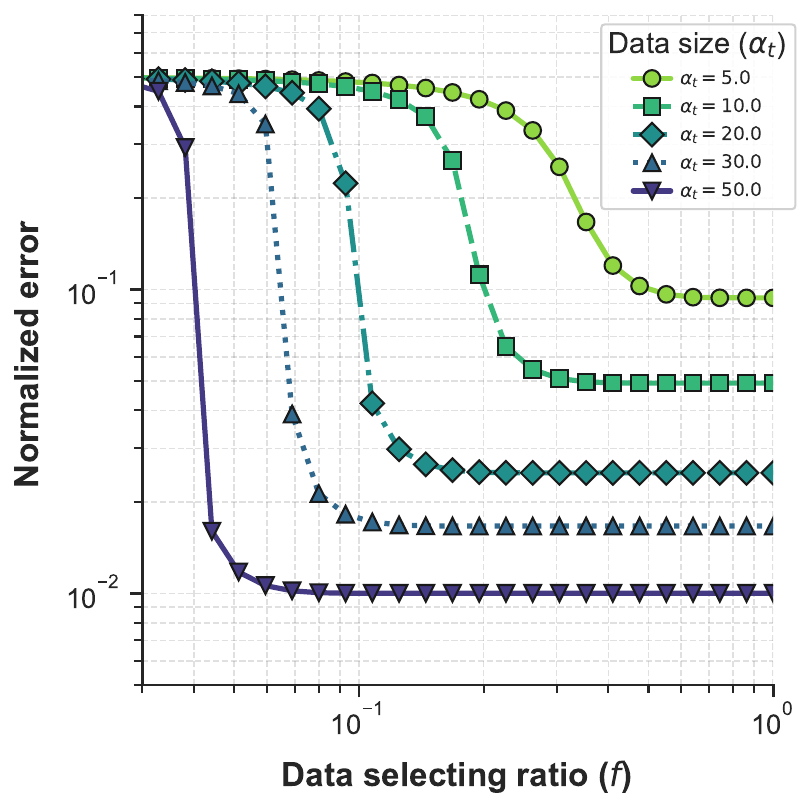}
        \caption{ $\theta=0^{\circ}$.}
        \label{fig:numerical_2}
    \end{subfigure}
    \begin{subfigure}[b]{0.49\columnwidth}
        \centering
        \includegraphics[width=\textwidth]{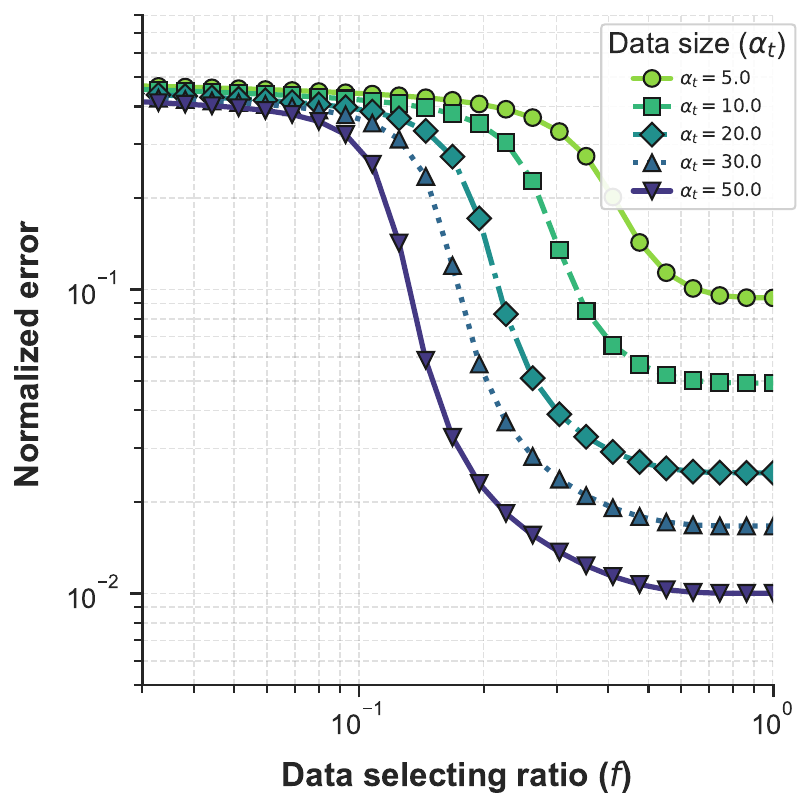}
        \caption{$\theta=20^{\circ}$.}
        \label{fig:numerical_4}
    \end{subfigure}
    \caption{The numerical experiments of $\mathcal{V}$-information aim to verify the impact of selecting hard samples on the performance of pre-trained foundation models in downstream tasks. Specifically, the correlation between the selection ratio $f$ and the fitting error is examined across different total data-to-parameter ratio $\alpha_t$ and overlap $\theta$. Moreover, as the angle $\theta$ increases, the model's fitting error increases as well. This indicates that using pre-trained models with domain distributions more closely aligned to the target task yields better downstream performance.}
    \label{fig:numer}
\end{figure}

\vspace{2pt}
\noindent
\textbf{Illustration through theoretical toy example. }We first validated this theoretically through the hard example selection toy example introduced in~\cref{sec:4.2}. By analyzing the downstream fitting error depicted in Figure~\ref{fig:numer}, we quantify the domain gap $\theta$ between the upstream pre-trained parameters and downstream tasks. This metric serves as a robust indicator for evaluating the efficacy of hard example selection strategies. Smaller $\theta$ values indicate greater relevance between the pre-trained model and downstream tasks. For example, parameters pre-trained on medical images exhibit stronger correlation with medical downstream tasks than those trained on ImageNet, resulting in smaller $\theta$ values. Our analysis further reveals that as $\theta$ increases, the error becomes more pronounced while maintaining the overall trend. This finding underscores the importance of selecting appropriate pre-training parameters for identifying hard examples in the OptiDEL model.

\vspace{2pt}
\noindent
\textbf{Validation on real-world medical datasets. }
To validate this theoretical finding on real medical data, we also use a pre-trained foundation model on ImageNet to select hard samples. While this method still performs reasonably well, we observe that its performance is slightly worse compared to using a model pre-trained on a more related task. This is because the ImageNet-pretrained model selects hard samples that are less accurate for the downstream task, leading to a decrease in performance. However, this strategy avoids the need to train a new model from scratch, reducing both computational overhead and time requirements. The results, detailed in \cref{tab:optidel_comparison}, showcase the performance of the DPT architectures as downstream frameworks. Although models pre-trained on more related tasks select hard samples more accurately and thus perform better, the ImageNet pre-trained model still achieves relatively similar results, further demonstrating the flexibility and efficiency of our method.


%% file: sec/6_conclution.tex
\section{Conclusion}
This paper transforms data-effective learning tasks into the optimization of $\mathcal{V}$-information, emphasizing that the increase in data does not always correlate with improved outcomes. It also points out that selecting the harder and generating more diverse samples allows for a smaller selection ratio without compromising performance as data volume increases. Guided by $\mathcal{V}$-information, we design the OptiDEL method, which outperforms traditional methods in multiple benchmarks, demonstrating practicality and efficiency in data-effective learning tasks. The paper provides theoretical and practical insights for designing more efficient data-effective learning methodologies, contributing to the advancement of medical foundation model pre-training.

%% file: sec/7_suppl.tex
\clearpage
\setcounter{page}{1}

\maketitlesupplementary
\appendix



\section{Optimizing V-information}
\subsection{Definitions of V-information}

Initially, we introduce the notion of $\mathcal{V}$-information, as formalized by \cite{Vinformation}.

\begin{definition}
\text{(Predictive Family). }Let $\Omega=\{f:\mathcal{X}\cup\{\varnothing\}\to\mathcal{P}(\mathcal{Y})\}$. A subset $\mathcal{V}\subseteq\Omega$ is called a predictive family if $\forall f \in \mathcal{V},  \forall P \in \mathrm{range}(f), \exists f^{\prime} \in \mathcal{V}, \text{s.t.} 
\, \forall x \in \mathcal{X}, f^{\prime}[x] = P, f^{\prime}[\varnothing] = P$.
\end{definition}

\begin{definition}
\text{(Predictive conditional $\mathcal{V}$-entropy).} Let $X, Y$ be two random variables taking values in $\mathcal{X} \times \mathcal{Y}$, and $\mathcal{V}$ be a predictive family. The predictive conditional $\mathcal{V}$-entropy is defined as:
\begin{align}
    H_{\mathcal{V}}(Y|X) &= \inf_{f \in \mathcal{V}} \mathbb{E}_{x,y \sim X,Y} \left[ -\log f[x](y) \right], \label{eq:app_conditional} \\
    H_{\mathcal{V}}(Y|\varnothing) &= \inf_{f \in \mathcal{V}} \mathbb{E}_{y \sim Y} \left[ -\log f[\varnothing](y) \right].
\end{align}
\end{definition}

\begin{definition}
\text{(Predictive $\mathcal{V}$-information).} Let $X, Y$ be two random variables taking values in $\mathcal{X} \times \mathcal{Y}$, and $\mathcal{V}$ be a predictive family. The predictive $\mathcal{V}$-information from $X$ to $Y$ is defined as:
\begin{equation}
    I_{\mathcal{V}}(X \to Y) = H_{\mathcal{V}}(Y|\varnothing) - H_{\mathcal{V}}(Y|X).
\end{equation}
\end{definition}

\citet{Vinformation} also introduce some properties of $\mathcal{V}$-information:
\begin{itemize}
    \item \textbf{Non-Negativity:}
    $$
        I_{\mathcal{V}}(X \to Y) \geq 0.
    $$
    \item \textbf{Independence:} If $X$ is independent of $Y$,
    $$
        I_{\mathcal{V}}(X \to Y) = 0.
    $$
\end{itemize}

\subsection{Sample Selection Guided by V-information}
In the context of data-effective learning, the predictive family $\mathcal{V}$ characterizes the self-supervised foundation model. Specifically, $X$ represents the input image space and $Y$ corresponds to the space of pseudo-labels. In contrastive learning, $Y$ represents whether two augmented views of an image belong to the same instance (positive pair) or different instances (negative pair), while in masked image modeling, $Y$ corresponds to the masked regions of the image that the model is trained to reconstruct.

Our objective is to increase the $\mathcal{V}$-information of a subset $D' \subseteq D$ through an algorithm $\Gamma$, where $D$ denotes the original dataset. The optimal selected subset $D'$ guided by $\mathcal{V}$-information can be formulated as:
\begin{align}
D'&= \mathop {arg\max} \limits_{D^{'}\subset D,|D^{'}|=f|D|} H_{\mathcal{V}}^{D^{'}}(Y|X). 
\end{align}
Our optimization goal for data selection strategy is to find a suitable algorithm that can increase the term $H_{\mathcal{V}}^{D^{'}}(Y|X)$.  
Based on the analysis in Section 4.2, the optimization target of the term $H_{\mathcal{V}}^{D^{'}}(Y|X)$ in the self-supervised foundation model's data-effective learning task can be represented as:
\begin{align}
D'&= \mathop {arg\max}\sum_{x_i, y_i \in D_{\text{d}}}  \log f_{D'}[x_i](y_i)
\end{align}
Here $D_{\text{d}}$ represents the downstream dataset. we present a detailed proof of Proposition 1 through a toy example to illustrate that selecting hard examples helps reduce the term $\sum_{x_i, y_i \in D_{\text{d}}}  \log f_{D'}[x_i](y_i)$ compared to random selection.  

\subsubsection{Details of the toy example.}

Let $\mathbf{\phi_I}$ be an $N$-dimensional target vector sampled from $\mathrm{Unif}(\mathbf{S}^{N-1}(\sqrt{N}))$, while $\mathbf{x}^\mu$ is sampled from an $N$-dimensional Gaussian distribution, and $y^{\mu}=\textbf{sign}(\mathbf{\phi_I}\cdot x)$ serves as the corresponding label. Together, they form the training dataset $D = \lbrace \mathbf{x}^\mu,\mathbf{y}^\mu \rbrace _{\mu = 1}^{P}$, used to predict $\mathbf{\phi_I}$.

By training on the full dataset $D$, we obtain a probe vector $\mathbf{\phi_{D}}$. For each sample $\mathbf{x}^\mu$, we define its margin as $m^\mu = \mathbf{\phi_D} \cdot (y^\mu\mathbf{x}^\mu)$. Based on these margins, we select a subset $S$ from $D$ at a pruning ratio $f$. From this subset, a new prediction vector $\mathbf{\phi_S}$ is trained from scratch. The effect of 
pruning is assessed by the alignment of $\mathbf{\phi_S}$ with the original target vector $\mathbf{\phi_I}$, measured as $R=\mathbf{\phi_I}\cdot\mathbf{\phi_S}/\|\mathbf{\phi_I}\|_{2}\|\mathbf{\phi_S}\|_{2}$. 

Here, the margin defined by $m^\mu =\mathbf{\phi_{D}}\cdot(y^\mu\mathbf{x}^\mu)$, serves as a metric for sample difficulty, where smaller margins indicate more challenging samples. The selection of hard samples involves identifying data points with the smallest margins. This process transforms the training data distribution from a Gaussian distribution to a narrower range concentrated around the sampling area's center, mathematically expressed as $p(z)= \frac{1}{\sqrt{2\pi}f}\exp(\frac{-z^2}{2})\Theta(\gamma-|z|)$, where $\tau = H^{-1}(\frac{1-f}{2})$. In this context, $\mathbf{\phi_S}$ is optimized by maximizing the margin defined as $m =\mathrm{min}_{\mu}\mathbf{\phi_S}\cdot(y^{\mu}\mathbf{x}^{\mu})$.

While \citet{sorscher} explored scenarios where the probe vector $\mathbf{\phi_{D}}$ does not perfectly match the teacher vector $\mathbf{\phi_{I}}$, our research provides a fresh perspective. We interpret $\mathbf{\phi_{D}}$ as representing the pre-trained foundation model's target function and $\mathbf{\phi_{I}}$ as the downstream task's target function. Within this framework, $\mathbf{\phi_{S}}$ models the foundation network trained on the selected subset. This perspective allows us to simulate how upstream pruning strategies influence the alignment between $\mathbf{\phi_{I}}$ and $\mathbf{\phi_{S}}$ in downstream tasks.



To quantify this relationship, we use the fitting error $e = \frac{arccos(R)}{\pi}$, which reflects the overlap between $\mathbf{\phi_{I}}$ and $\mathbf{\phi_{S}}$. Additionally, we define the deviation between $\mathbf{\phi_{D}}$ and $\mathbf{\phi_{I}}$ by angle $\theta$. We further consider the role of data-to-parameter ratios in this process. Let $\alpha_t=P/N$ represents the initial ratio of data points to model parameters, and $\alpha_s = f\alpha_t$ is the adjusted ratio post-pruning. In self-supervised pre-training tasks, $\alpha_t$ is fixed, representing the full utilization of training data for a given model architecture.

Our analysis focuses on how difficulty-based pruning impacts downstream performance, as measured by the fitting error $e$, and examines how data volume influences pruning efficacy under a fixed number of training parameters. This approach highlights the interplay between pruning strategies and downstream model performance, providing insights into optimizing the selection of training examples for foundation models.

\citet{sorscher} built upon Elizabeth Gardner's approach \cite{space} to investigate how the fitting performance $R$ of the pruned subset varies with $\alpha_t$. Specifically, it has been demonstrated that $R$ can be obtained as the solution to the following system of equations:
\begin{equation}
\left\{
\begin{aligned}
& \frac{R - \rho \cos\theta}{\sin^2\theta} = \frac{\alpha_s}{\pi \Psi} \langle \int_{-\infty}^{m} dx \, \exp\left(-\frac{G(x,z)}{2\Psi^2}\right) (m - x) \rangle_z \\
& \frac{\sin^2\theta-\rho^2 - R^2 + 2\rho R \cos\theta}{\sin^2\theta} = 2\alpha_s \langle \int_{-\infty}^{\kappa} (m - x)^2 dx \\
& \frac{e^{-\frac{(x-\rho z)^2}{2(1-\rho^2)}}}{\sqrt{2\pi(1-\rho^2)}}   
H\left(\frac{F(x,z)}{\sqrt{1-\rho^2}\Psi}\right) \rangle_z \\
& \frac{\rho - R\cos\theta}{\sin^2\theta} = 2\alpha_s \langle \int_{-\infty}^{m} dx \, \frac{e^{-\frac{(x-\rho z)^2}{2(1-\rho^2)}}}{\sqrt{2\pi(1-\rho^2)}} (m - x) \\ 
& H\left(\frac{F(x,z)}{\sqrt{1-\rho^2}\Psi}\right) \frac{z - \rho x}{1-\rho^2} \rangle_z
\end{aligned}
\right.
\label{eq:hard}
\end{equation}
where the auxiliary functions are defined as:
\begin{equation}
\left\{
\begin{aligned}
& \Psi = \sqrt{\sin^2\theta - R^2 - \rho^2 + 2\rho R \cos\theta} \\
& F(x,z) = z(\rho R - \cos\theta) - x(R - \rho\cos\theta) \\
& G(x,z) = z^{2}(\rho^{2} + \cos^{2}\theta - 2\rho R\cos\theta) \\
& \quad + 2xz(R\cos\theta - \rho) + x^{2}\sin^{2}\theta
\end{aligned}
\right.
\end{equation}
and the order parameter $\rho = \mathbf{\phi_{D}} \cdot \mathbf{\phi_{S}}$ quantifies the alignment between the target function of the pre-trained foundation model and the foundation network trained on the selected subset.

In our framework, by incorporating the constraint $\frac{\alpha_s}{f}\equiv c$ into ~\citet{sorscher}'s analysis, we reformulate Equation \eqref{eq:hard} as:
\begin{equation}
\left\{
\begin{aligned}
& \frac{R - \rho \cos\theta}{\sin^2\theta} = \frac{cf}{\pi \Psi} \langle \int_{-\infty}^{m} dx \, \exp\left(-\frac{G(x,z)}{2\Psi^2}\right) (m - x) \rangle_z \\
& \frac{\sin^2\theta-\rho^2 - R^2 + 2\rho R \cos\theta}{\sin^2\theta} = 2cf \langle \int_{-\infty}^{\kappa} (m - x)^2 dx \\
& \frac{e^{-\frac{(x-\rho z)^2}{2(1-\rho^2)}}}{\sqrt{2\pi(1-\rho^2)}}   
H\left(\frac{F(x,z)}{\sqrt{1-\rho^2}\Psi}\right) \rangle_z \\
& \frac{\rho - R\cos\theta}{\sin^2\theta} = 2cf \langle \int_{-\infty}^{m} dx \, \frac{e^{-\frac{(x-\rho z)^2}{2(1-\rho^2)}}}{\sqrt{2\pi(1-\rho^2)}} (m - x) \\ 
& H\left(\frac{F(x,z)}{\sqrt{1-\rho^2}\Psi}\right) \frac{z - \rho x}{1-\rho^2} \rangle_z
\end{aligned}
\right.
\label{eq:hard}
\end{equation}

Through numerical calculations, we obtain $R$ and the corresponding fitting error $e$ under various pre-training data volumes $\alpha_t$ and pruning ratios $f$. Our results reveal that when $\alpha_t$ is fixed at a large value $c$, $e$ rapidly decreases and approaches zero at small $f$ values. This indicates that selecting hard examples effectively identifies the most informative data segments and achieves comparable performance to models trained on the complete dataset, so we have:
\begin{multline}
\frac{1}{|D_{\text{d}}|} \sum_{(x_i, y_i) \in D_{\text{d}}} \mathbb{I}\big[\arg\max f_{D_\text{hard}}[x_i] = y_i\big] \\
- \frac{1}{|D_{\text{d}}|} \sum_{(x_i, y_i) \in D_{\text{d}}} \mathbb{I}\big[\arg\max f_{D_\text{random}}[x_i] = y_i\big] > 0
\end{multline}

Considering the close connection between model performance and $f[x](y)$ which is a probability measure on $y \in \mathcal{Y}$ given side information $x$, we can derive the following inequality:
\begin{equation}
\sum_{x_i, y_i \in D_{\text{d}}}  \left(\log f_{D_\text{hard}}^{}[x_i](y_i)-\log f_{D_\text{random}}[x_i](y_i)\right )>0
\end{equation} 
This inequality aligns precisely with the form presented in Proposition 1, thereby providing a theoretical foundation for the empirical advantage of hard example selection in model training. 

\section{Implementation Details}

All experiments are performed on 4*NVIDIA RTX 3090 GPUs with 24GB memory each.

\begin{table*}[t]
    \centering
    \vspace{1mm}
    \caption{Comparison of OptiDEL with state-of-the-art baseline methods using MAE+FCNHead architecture. To evaluate the performance of our OptiDEL method, we pre-train foundation models using 5\%, 10\%, 25\%, and 50\% of the total data volume and test them on eight downstream datasets using EL2N, RDED, MedDEL, and OptiDEL. OptiDEL surpasses other methods in nearly all datasets, demonstrating its stability and efficiency.}
    \renewcommand{\arraystretch}{1.3}  
    \setlength{\tabcolsep}{5pt}  
    \resizebox{\textwidth}{!}{  
        \begin{tabular}{l| c | c c c c | c c c c | c c c c | c c c c}
            \toprule
            \multirow{2}{*}{\diagbox{\textbf{Dataset}}{\textbf{mIoU}}} 
            & \textbf{All Data} 
            & \multicolumn{4}{c}{\textbf{5\% storage volume}} 
            & \multicolumn{4}{|c}{\textbf{10\% storage volume}} 
            & \multicolumn{4}{|c}{\textbf{25\% storage volume}} 
            & \multicolumn{4}{|c}{\textbf{50\% storage volume}} \\
            \cmidrule(lr){2-2} \cmidrule(lr){3-6} \cmidrule(lr){7-10} \cmidrule(lr){11-14} \cmidrule(lr){15-18}
            & \textbf{Baseline} 
            & \textbf{EL2N} & \textbf{RDER} & \textbf{MedDEL} & \textbf{OptiDEL} 
            & \textbf{EL2N} & \textbf{RDER} & \textbf{MedDEL} & \textbf{OptiDEL} 
            & \textbf{EL2N} & \textbf{RDER} & \textbf{MedDEL} & \textbf{OptiDEL} 
            & \textbf{EL2N} & \textbf{RDER} & \textbf{MedDEL} & \textbf{OptiDEL} \\
            \midrule
            Kvasir-Instrument & 84.76 & 83.24 & 84.25 & 83.89 & \textbf{86.35} & 83.89 & 84.44 & 82.99 & \textbf{85.41} & 83.29 & 85.57 & 84.39 & \textbf{86.11} & 83.80 & 85.05 & 85.51 & \textbf{86.98} \\
            Kvasir-SEG & 62.01 & 60.21 & 62.87 & 62.09 & \textbf{64.74} & 58.90 & 60.88 & 59.04 & \textbf{63.22} & 60.09 & 62.24 & 61.57 & \textbf{63.95} & 60.58 & 62.21 & 61.44 & \textbf{62.89} \\
            ImageCLEFmed & 58.09 & 56.54 & 58.44 & 58.19 & \textbf{59.71} & 57.07 & \textbf{58.93} & 58.19 & 57.95 & 58.67 & 58.44 & 56.15 & \textbf{58.76} & 54.84 & 58.11 & 55.51 & \textbf{60.78} \\
            ETIS & 63.51 & 64.18 & 65.26 & 65.09 & \textbf{70.38} & 62.04 & 63.53 & 63.13 & \textbf{64.77} & 64.91 & 67.20 & 65.62 & \textbf{68.48} & 63.27 & 64.83 & 66.75 & \textbf{68.28} \\
            PolypGen2021 & 46.07 & 45.18 & 46.30 & 44.28 & \textbf{47.97} & 37.12 & 40.60 & 38.70 & \textbf{43.51} & 43.41 & 44.93 & 44.79 & \textbf{47.96} & 44.12 & \textbf{47.24} & 44.04 & 47.01 \\
            CVC-300 & 83.90 & 84.02 & 84.99 & 84.76 & \textbf{86.97} & 82.37 & 85.24 & 83.12 & \textbf{86.04} & 85.41 & 86.11 & 84.53 & \textbf{87.63} & 85.83 & 86.22 & \textbf{87.07} & 86.74 \\
            CVC-ClinicDB & 80.56 & 79.94 & 79.94 & 79.91 & \textbf{80.92} & 77.42 & 76.44 & 77.36 & \textbf{81.63} & 78.83 & 80.93 & 78.39 & \textbf{81.58} & 78.75 & \textbf{82.75} & 80.42 & 81.34 \\
            CVC-ColonDB & 77.46 & 75.12 & 77.90 & 77.54 & \textbf{78.67} & 74.68 & 74.10 & 74.10 & \textbf{75.15} & 70.49 & 75.27 & 70.80 & \textbf{76.43} & 72.24 & 75.64 & 75.08 & \textbf{77.36} \\
            \bottomrule
        \end{tabular}
    }  
        \label{app_tab1}
\end{table*}

\begin{table*}[t]
    \centering
    \caption{Analysis of OptiDEL across different foundation models using the MAE+FCNHead architecture. We compare two variants: OptiDEL(ImageNet), which employs MAE models pre-trained on the ImageNet dataset, and OptiDEL which employs MAE models pre-trained on LDPolypVideo~\cite{LDVideo} and Hyper-Kvasir~\cite{HyperKvasir}. Our results indicate that OptiDEL(ImageNet), despite showing marginally lower performance than its medical dataset-pretrained counterpart, maintains impressive effectiveness and frequently surpasses the performance of models trained on the complete dataset.}
    \vspace{-5pt}
    \renewcommand{\arraystretch}{1.2}  
    \setlength{\tabcolsep}{5pt}  
    \resizebox{\textwidth}{!}{  
        \begin{tabular}{l| c | c c | c c | c c | c c}
            \toprule
            \multirow{2}{*}{\diagbox{\textbf{Dataset}}{\textbf{mIoU}}} 
            & \textbf{All Data} 
            & \multicolumn{2}{c}{\textbf{5\% storage volume}} 
            & \multicolumn{2}{|c}{\textbf{10\% storage volume}} 
            & \multicolumn{2}{|c}{\textbf{25\% storage volume}} 
            & \multicolumn{2}{|c}{\textbf{50\% storage volume}} \\
            \cmidrule(lr){2-2} \cmidrule(lr){3-4} \cmidrule(lr){5-6} \cmidrule(lr){7-8} \cmidrule(lr){9-10}
            & \textbf{Baseline} 
            & \textbf{OptiDEL (Imagenet)} & \textbf{OptiDEL} 
            & \textbf{OptiDEL (Imagenet)} & \textbf{OptiDEL}  
            & \textbf{OptiDEL (Imagenet)} & \textbf{OptiDEL}  
            & \textbf{OptiDEL (Imagenet)} & \textbf{OptiDEL}  \\
            \midrule
            Kvasir-Instrument & 84.76 & 84.09 & \textbf{86.35} & 85.12 & \textbf{85.41} & 85.60 & \textbf{86.11} & 86.17 & \textbf{86.98}\\
            Kvasir-SEG & 62.01 & 61.65 & \textbf{64.74} & 58.09 & \textbf{63.22} & 60.69 & \textbf{63.95} & 61.54 & \textbf{62.89} \\
            ImageCLEFmed & 58.09 & 57.35 & \textbf{59.71} & 57.82 & \textbf{57.95} & 58.44 & \textbf{58.76} & 58.98 & \textbf{60.78} \\
            ETIS & 63.51 & 64.59 & \textbf{70.38} & 63.96 & \textbf{64.77} & 67.29 & \textbf{68.48} & 66.10 & \textbf{68.28} \\
            PolypGen2021 & 46.07 & 47.18 & \textbf{47.97} & 40.68 & \textbf{43.51} & 45.83 & \textbf{47.96} & 46.60 & \textbf{47.01}\\
            CVC-300 & 83.90 & 85.74 & \textbf{86.97} & 85.78 & \textbf{86.04} & 86.37 & \textbf{87.63} & 85.64 & \textbf{86.74}\\
            CVC-ClinicDB & 80.56 & 80.85 & \textbf{80.92} & 80.84 & \textbf{81.63} & 80.96 & \textbf{81.58} & 81.59 & \textbf{81.34} \\
            CVC-ColonDB & 77.46 & 77.82 & \textbf{78.67} & 74.68 & \textbf{75.15} & 73.53 & \textbf{76.43} & 76.71 & \textbf{77.36} \\
            \bottomrule
        \end{tabular}
    }  
        \label{app_tab2}
\end{table*}

\subsection{Medical Task Settings}

\textbf{Pre-training foundation models to select hard samples. }To efficiently identify hard samples in real-world medical images, we leverage pre-trained foundation models to score each sample, where samples with higher loss values are classified as hard samples. For evaluation, we employ Masked AutoEncoder models (MAE) pre-trained on ImageNet~\cite{imagenet} and two medical-specific datasets: LDPolypVideo~\cite{LDVideo} and Hyper-Kvasir~\cite{HyperKvasir}.

\vspace{2pt}
\noindent
\textbf{Network architectures. }For model training, we adopt a comprehensive two-phase approach. In the pre-training phase, we utilize Masked Auto Encoder (MAE-ViT-Base)\cite{He_2022_CVPR} for self-supervised foundation model training, with a batch size of 16 over 100 epochs. We employ the AdamW optimizer\cite{2017Decoupled} with a cosine learning rate schedule~\cite{gotmare2018closer}, maintaining a peak learning rate of $1.5 \times 10^{-4}$. For downstream tasks, we evaluate two architectures under the MMSegmentation framework~\cite{mmseg2020}: MAE with FCNHead~\cite{FCN} and Dense Prediction Transformer (DPT)~\cite{DPT}. The downstream training runs for 20,000 iterations using AdamW optimizer with a maximum learning rate of $6 \times 10^{-5}$.
\subsection{Numerical Calculation Settings}

We prune each synthetic dataset in all numerical experiments to retain a fraction $f$ of the smallest-margin examples. We fix the parameter number at $N=30$ and determine the size of the sub-dataset as $P=fN\alpha_t$, where $\alpha_t$ represents the ratio of data to parameters that we specify. The fraction $f$ is sampled logarithmically from $0.01$ to $1$. We use a standard quadratic programming algorithm to optimize the perceptrons and find the maximum-margin separating solution. Results are averaged over 50 independent draws of the origin and selected examples.

For the angle deviation experiment, we obtain numerical solutions at different angles by varying $\theta$. In the data redundancy experiment, to simulate redundancy for $n\times$, we randomly sample $\frac{P}{n}$ points and replicate them $n-1$ times.

\section{Additional Experimental Results}


\begin{figure*}[t]
    \centering
    \begin{subfigure}[b]{0.33\textwidth}
        \centering
        \includegraphics[width=\textwidth]{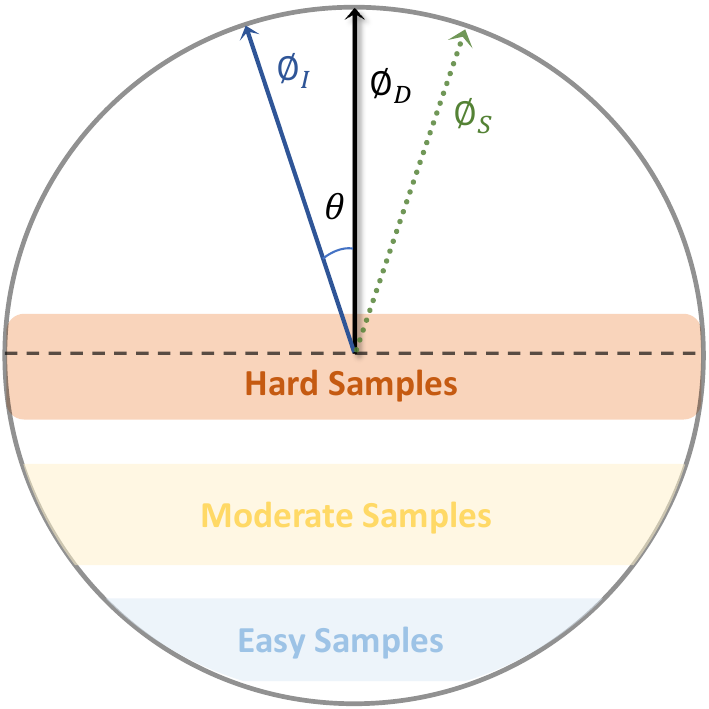}
        \caption{Types of samples.}
        \label{fig:numerical_3}
    \end{subfigure}
    \begin{subfigure}[b]{0.33\textwidth}
        \centering
        \includegraphics[width=\textwidth]{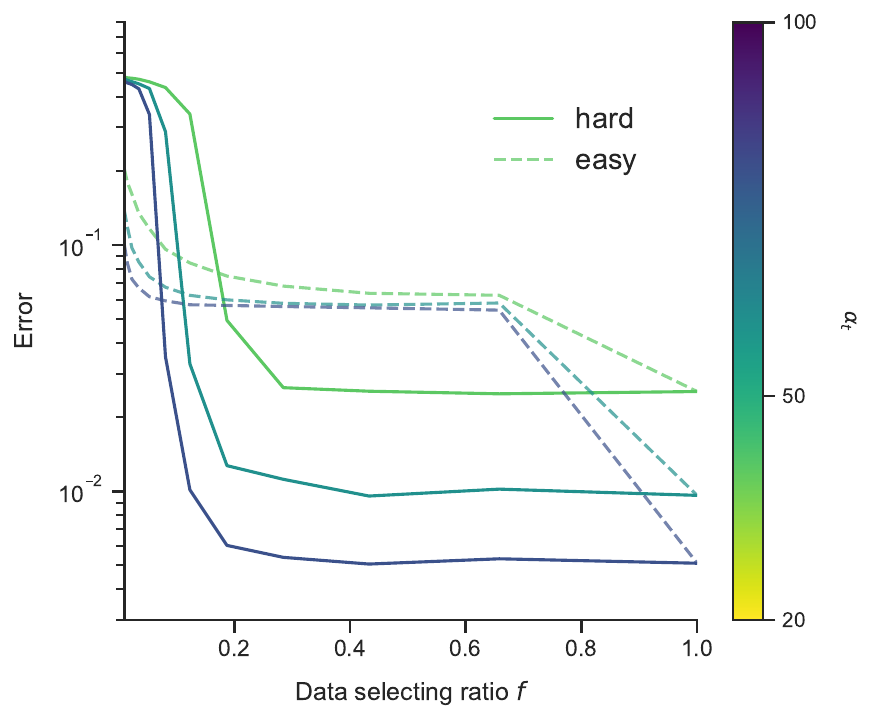}
        \caption{Selecting hard samples vs easy samples.}
        \label{fig:numerical_4}
    \end{subfigure}
    \begin{subfigure}[b]{0.33\textwidth}
        \centering
        \includegraphics[width=\textwidth]{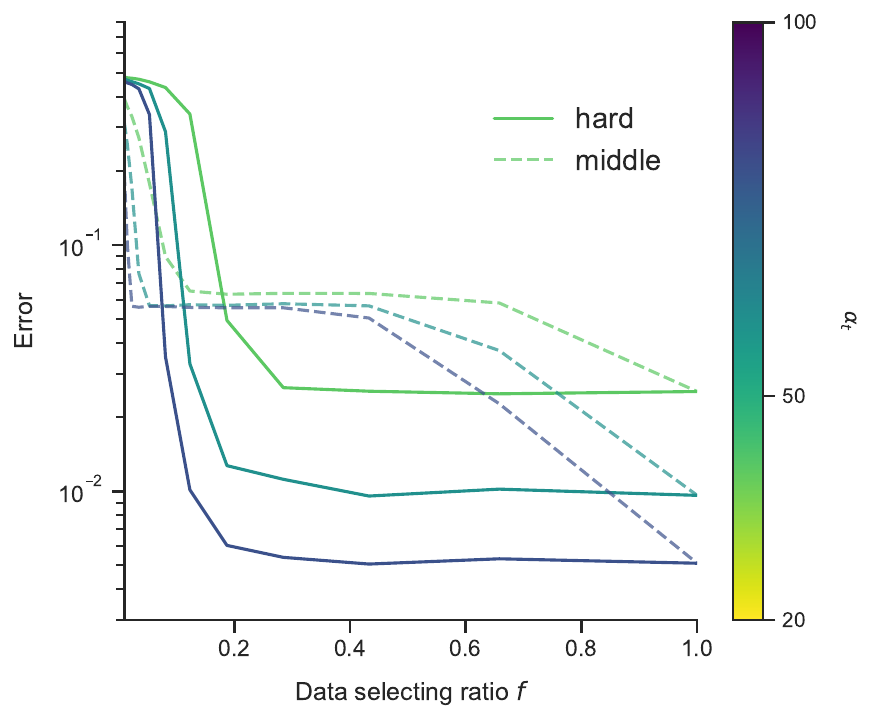}
        \caption{Selecting hard samples vs moderate samples.}
        \label{fig:numerical_4}
    \end{subfigure}
    \caption{The numerical experiments aim to verify the relationship between fitting error $e$ and pruning ratio $f$ under different difficulty-based sample methods of pre-training foundation models in downstream tasks. The solid line represents the selection of hard samples, while the dotted lines correspond to random selection, easy sample selection, and moderate sample selection in the three plots, respectively. It can be observed that when the total data size is large, selecting difficult samples generally yields better performance.}
    \label{fig:numerical4}
\end{figure*}
\begin{figure*}[ht!]
    \centering
    \begin{subfigure}[b]{0.24\textwidth}
        \centering
        \includegraphics[width=\textwidth]{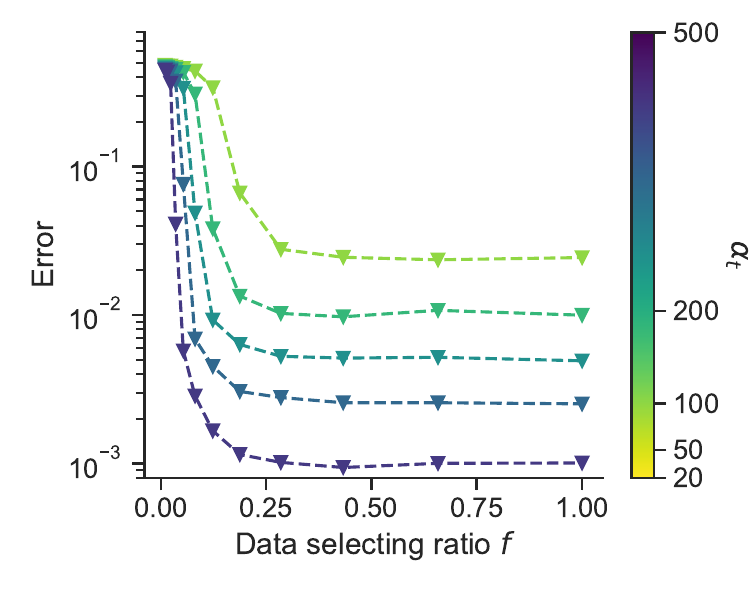}
        \caption{No data redundancy.}
        \label{fig:Numerical_1}
    \end{subfigure}
    \begin{subfigure}[b]{0.24\textwidth}
        \centering
        \includegraphics[width=\textwidth]{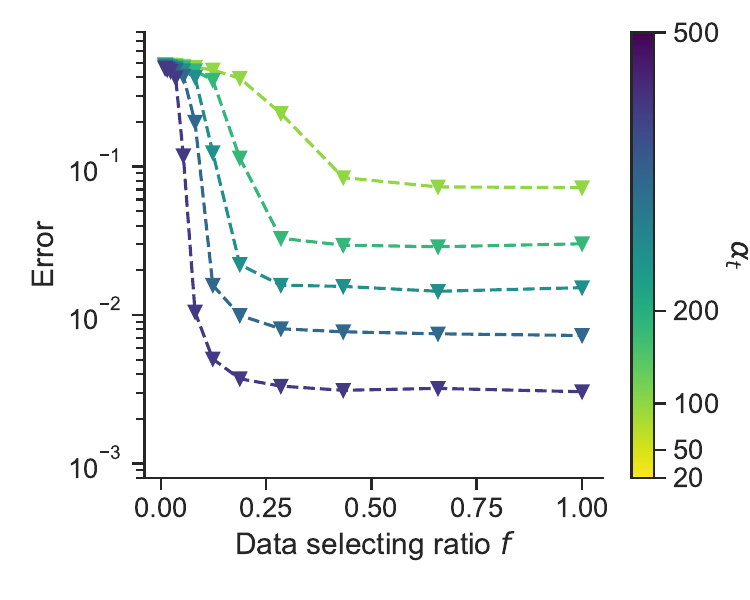}
        \caption{3x data redundancy.}
        \label{fig:Numerical_2}
    \end{subfigure}
    \begin{subfigure}[b]{0.24\textwidth}
        \centering
        \includegraphics[width=\textwidth]{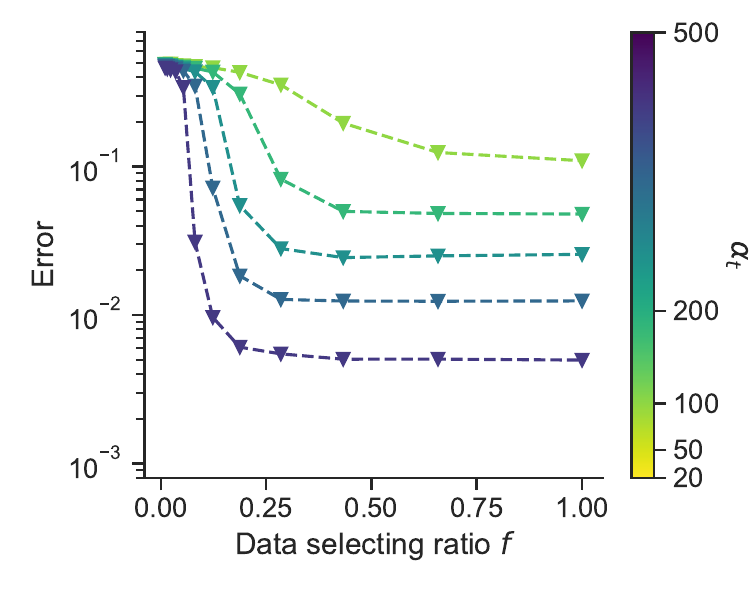}
        \caption{5x data redundancy.}
        \label{fig:Numerical_3}
    \end{subfigure}
    \begin{subfigure}[b]{0.24\textwidth}
        \centering
        \includegraphics[width=\textwidth]{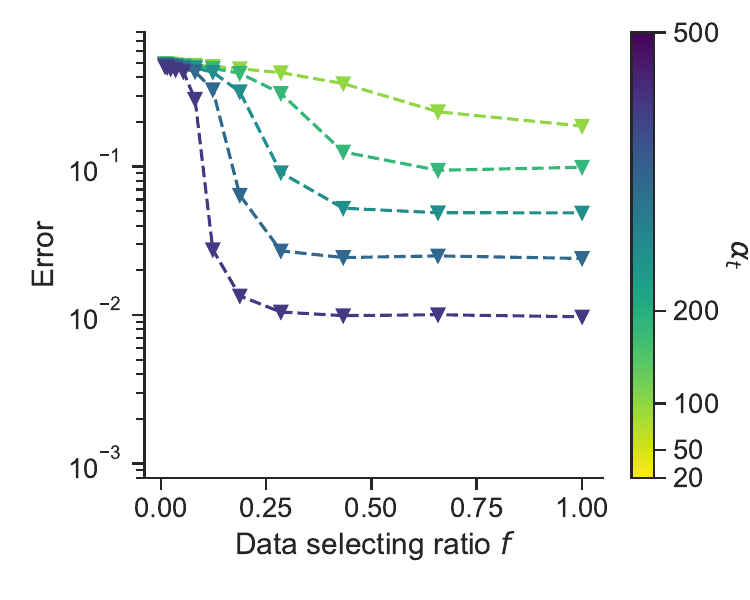}
        \caption{10x data redundancy.}
        \label{fig:Numerical_4}
    \end{subfigure}
    \caption{The numerical experiments of $\mathcal{V}$-information aim to verify the impact of data redundancy on the robustness of pre-trained foundation models. Specifically, the correlation between the fitting error (defined as the discrepancy between the target function of the downstream task $\phi_I$ and the foundation model pre-trained with distilled data $\phi_S$) and the data selection ratio $f$ is examined across different levels of data redundancy (0x, 3x, 5x, 10x), here we set $\theta=10^{\circ}$. The results demonstrate that as data redundancy increases,  selecting a small proportion of the dataset still yields performance comparable to using the full dataset. This indicates the potential in real-world training scenarios to distill more effective datasets from larger data pools with redundant data.}
    \label{fig:Numerical}
\end{figure*}



\subsection{Ablation of Downstream Model Structure}
We evaluate the performance of OptiDEL against related methods, as well as variants of OptiDEL with different pre-trained parameters, using the MAE+FCNHead architecture as the downstream framework. The results, presented in \cref{app_tab1} and \cref{app_tab2}, demonstrate OptiDEL's superior performance across diverse downstream frameworks, highlighting its robustness and adaptability.

\subsection{More Numerical Calculation Results}

\noindent
\textbf{More data selection strategies. }In our experiments, we focus on selecting hard samples. Prior research has also explored the benefits of incorporating easy and moderately challenging examples into the selection process~\cite{sorscher,moderate}. While innovative strategies based on sample difficulty continue to emerge~\cite{ccs,d2}, many of these methods introduce additional parameters, leading to higher computational overhead when applied to large datasets. 

In this section, we focus on comparing the performance of selecting hard, easy, and moderately challenging samples within our toy example. Through a detailed analysis, we demonstrate the rationale behind ultimately choosing hard samples for our approach, highlighting their contribution to achieving superior performance.

Specifically, we analyze the model's error rate $e$ across different data selection ratios $f$ by comparing four distinct sample selection approaches: difficult samples, easy samples (characterized by the largest margin), moderate samples (exhibiting intermediate margin values), and randomly selected samples. Throughout our experiments, we maintain a constant total data volume of $ \alpha_t=c $ and present our findings through comprehensive numerical analysis.

Figure~\ref{fig:numerical4} clearly illustrates the strong correlation between experimental outcomes and the total volume of data. When the data volume is large and
$f$ is not too small, selecting hard samples consistently outperforms random selection, as well as the selection of easy or moderate samples. This observation supports our focus on hard examples in the main text, particularly in scenarios with abundant pre-training data. However, when the data volume is small, the effectiveness of selecting hard samples diminishes compared to the other three methods, especially under high pruning rates. This result aligns with the experimental findings of \citet{sorscher,moderate} on real-world image datasets.

It is worth emphasizing that the data in our toy example consists solely of vectors. In real-world applications, data is often more complex and must account for additional considerations such as diversity and class imbalance. Despite this, our conclusions offer valuable insights and practical guidance for sample selection based on data difficulty.

\vspace{2pt}
\noindent
\textbf{The impact of data redundancy. }To further explore the performance of selecting the hardest samples under data redundancy which is common in the medical field, we maintain the total volume of data constant while replicating the dataset 3x, 5x, and 10x times in Figure~\ref{fig:Numerical_2}, Figure~\ref{fig:Numerical_3}, and Figure~\ref{fig:Numerical_4}, and then perform numerical calculations in the toy example. The results reveal that as data redundancy increases, the rate of performance improvement of the foundation model slows down with the increase in pre-training data volume. However, even with a large original data volume, it is still possible to achieve higher performance by selecting a smaller proportion of the dataset. This suggests that in real-world large dataset training tasks, it is still feasible to distill a smaller, more effective dataset from a larger pool to enhance performance.